%% file: main.tex
\documentclass{ws-ijalp}
\usepackage[utf8]{inputenc}
\usepackage[T5]{fontenc}
\usepackage{hyperref}
\usepackage{tabularx}
\usepackage{multirow}
\usepackage{amssymb}
\usepackage{xcolor}
\usepackage{pgfplots}
\usepackage{amsmath}
\usepackage{float}
\usepackage{tikz}
\usepackage{array,booktabs}
\usepackage{graphicx}
\usepackage{caption}
\usepackage[para,online,flushleft]{threeparttable}

\begin{document}

\markboth{Pre-print}{Metadata Integration for Spam Reviews Detection on Vietnamese E-commerce Websites}

\catchline{}{}{}{}{}

\title{Metadata Integration for Spam Reviews Detection on Vietnamese E-commerce Websites}

\author{Co Van Dinh}

\address{Faculty of Information Science and Engineering, University of Information Technology, Ho Chi Minh City, Vietnam \\ Vietnam National University, Ho Chi Minh City, Vietnam
\\
19521293@gm.uit.edu.vn}

\author{Son T. Luu\footnote{Corresponding author}}

\address{Faculty of Information Science and Engineering, University of Information Technology, Ho Chi Minh City, Vietnam \\ Vietnam National University, Ho Chi Minh City, Vietnam
\\
sonlt@uit.edu.vn}

\maketitle

\begin{history}
\received{(Day Month Year)}
\revised{(Day Month Year)}
\end{history}

\begin{abstract}
The problem of detecting spam reviews (opinions) has received significant attention in recent years, especially with the rapid development of e-commerce. Spam reviews are often classified based on comment content, but in some cases, it is insufficient for models to accurately determine the review label. In this work, we introduce the ViSpamReviews v2 dataset, which includes metadata of reviews with the objective of integrating supplementary attributes for spam review classification. We propose a novel approach to simultaneously integrate both textual and categorical attributes into the classification model. In our experiments, the product category proved effective when combined with deep neural network (DNN) models, while text features performed well on both DNN models and the model achieved state-of-the-art performance in the problem of detecting spam reviews on Vietnamese e-commerce websites, namely PhoBERT. Specifically, the PhoBERT model achieves the highest accuracy when combined with product description features generated from the SPhoBert model, which is the combination of PhoBERT and SentenceBERT. Using the macro-averaged F1 score, the task of classifying spam reviews achieved 87.22\% (an increase of 1.64\% compared to the baseline), while the task of identifying the type of spam reviews achieved an accuracy of 73.49\% (an increase of 1.93\% compared to the baseline).
\end{abstract}


\input{introduction.tex}

\input{relatedworks.tex}
\input{method.tex}
\input{results.tex}
\input{conclusion.tex}

\section*{Acknowledgments}
This research was supported by The VNUHCM-University of Information Technology's Scientific Research Support Fund

\bibliographystyle{ws-ijalp}
\bibliography{references}

\end{document}

%% file: introduction.tex
\section{Introduction}
\label{intro}

In recent years, e-commerce in Vietnam has witnessed rapid and strong development. Many e-commerce platforms have emerged and operated in Vietnam, changing the way consumers shop. The number of consumers who shop online in Vietnam has significantly increased in recent years. According to statistics \cite{eco2022}, now over half of the Vietnamese population participates in online shopping on e-commerce websites. This is due to the convenience it brings to users, as they do not need to visit the stores directly but can simply make a few clicks on their digital devices, and their orders will be delivered to their doorsteps. However, this has also led to some users being deceived into buying fake or low-quality items that do not match the descriptions. As a result, this is negatively impacting the experiences of consumers when using online shopping services.

The shopping behavior of customers is significantly influenced by the reviews provided by customers who have purchased the product previously. Customers tend to look at reviews and feedback about a product to decide whether or not to purchase it. With the vast development of e-commerce platforms, the opinion and experience of users about the product or company is the key source of information for the customers \cite{NI2024101358}. Catching up with this behavior, some sellers create fake reviews that do not reflect the actual quality of the product, or some other users intentionally create negative reviews to damage the reputation of the store \cite{jindal2007review,YOU2011510,RAO2021115742}. In addition, the e-commerce platforms encourage users to review products to receive vouchers, which results in reviews that are completely irrelevant to the product. These reviews are called spam reviews because it is not useful for users. Moreover, they can easily be misled into buying low-quality products that do not match the description. Therefore, removing these spam reviews helps users avoid purchasing low-quality products and provides a better shopping experience.

In \cite{vispamreviews}, the authors introduced the ViSpamReviews dataset for detecting spam reviews based on user comments. However, through experimentation and error analysis, we found that using only user comments was not enough to accurately determine whether a review was spam or not. This is because some comments do not provide sufficient information related to the product, which makes it difficult for the model to predict the label of the review and leads to inaccurate predictions. Therefore, in this work, we released the ViSpamReviews v2 dataset, which was constructed on the previous dataset. In this version, in addition to comments and star ratings, we also collected some product-related information including product category, product name, product description, number of sold, and number of reviews. This information, also known as metadata, can be integrated into the classification model to improve its accuracy compared to using only user comments to detect spam reviews in online e-commerce platforms.

Our main contributions to this research are as follows:

\begin{itemize}

    \item Firstly, we introduce the ViSpamReviews v2 dataset, which includes user reviews and metadata for the task of detecting spam reviews on Vietnamese e-commerce websites. In this new dataset, we add the metadata for each comment from the ViSpamReview \cite{vispamreviews} without changing its annotated label.  
    
    \item Secondly, we propose a method to integrate metadata for the problem of spam review classification based on combining categorical \cite{kim-etal-2019-categorical} and textual attributes \cite{reimers-2019-sentence-bert} to deep neural networks for the spam detection task. 

    \item Finally, we perform multiple experiments to evaluate the results of the metadata combination methods on various classification methods.
    
\end{itemize}

The paper is structured as follows. Section \ref{intro}, introduces our research as presented above. Section \ref{related_works}, surveys relevant research on the problem of online spam detection and the methods for integrating metadata into the classification models. Section \ref{method}, introduces our ViSpamReviews v2 dataset, analyzes the metadata in the new dataset, and proposes a method for integrating product category and description into our problem. Section \ref{result}, presents the steps of data preprocessing, implementation of experimental parameters, the results of our experiments, and analyzes the errors for each method. Finally, Section \ref{conclusion} concludes our research.

%% file: relatedworks.tex
\section{Related Works}
\label{related_works}

Jindal and Liu introduced the task of Spam Reviews \cite{jindal2007review,jindal2008opinion} to minimize the problem of spreading unreliable opinions about products or services and preserve the trustworthiness of potential customers on e-commerce platforms. With the vast development of e-commerce website platforms, the spam reviews task plays an important role in helping customers find reliable opinions about a product as a reference in shopping online. Besides, the spam reviews task also helps the shop owner access valuable reviews to improve the quality of products and services and attract the trustworthiness of current and potential customers. The spam reviews task is categorized as the text classification task \cite{jindal2007review}. As shown in \cite{9140745,huynh-etal-2020-simple,SHEN2023101283}, deep neural models (DNN) such as LSTM \cite{hochreiter1997long}, GRU \cite{cho-etal-2014-learning}, and TextCNN \cite{kim-2014-convolutional} outperform the text classification task for Vietnamese languages in comparison with the traditional models such as Logistic Regression and SVM. In addition, according to the benchmark evaluation works in \cite{nguyen-etal-2022-smtce}, the BERTology models \cite{rogers-etal-2020-primer} fine-tuned on Vietnamese social text also bring state-of-the-art results. Therefore, we use deep neural networks and BERTology approaches to construct the classification model for the spam detection task. Besides, the SentenceBERT \cite{reimers-2019-sentence-bert} is a powerful model to retrieve the potential information from sentences in texts such as users' reviews and descriptions of products. 

In Vietnamese, the ViSpamReviews \cite{vispamreviews} is a large-scale dataset for spam review detection with 19K reviews from users on Vietnamese e-commerce platforms. The dataset is manually annotated by humans and conducts a ground-truth process by major voting. The authors also provide several baseline methodologies based on deep neural network and transformer-based language models and obtain optimistic results. Also, Najadat et al. \cite{10.1145/3476115} provide a dataset for online spam review detection on social networks in Arabic. However, according to \cite{10.1145/3522575}, with the vast increment of customer reviews on e-commerce and the requirement of enterpriser to find valuable information about the products from customers' reviews, it is necessary to integrate the extra information about the product features with the texts of the reviews from users to improve the models. To extract the metadata for spam detection, authors in \cite{10.1145/2783258.2783370} use the behaviors (review contents, rating, and product information) combined with text mining based on review texts from users and graph-based methodology to model the relationship between user, reviews, and products. However, with the vast information on e-commerce platforms, the graph for representing the relationship between users and products will be enormous and complicated, which challenges computing resources to perform calculations. Hence, the potential approaches are based on the content of both users and products on the e-commerce websites, including the review texts, categories of products, names of products and brands, and product description texts. In \cite{kim-etal-2019-categorical}, the authors integrate the categorical information of products with the review text of users and showed an improvement in the Review Sentiment Tasks performance. Additionally, Cao et al. \cite{10.1145/3522575} extract the features of products by semantic-based methods including LDA clustering for feature words and pos-tag extraction to extract the linguistic information from review texts.  

The metadata improves the text classification model by providing extra information and semantic context. According to \cite{SATIABUDHI2021101048}, there are two main approaches to detecting spam reviews from customers, including content-based and behavior-based. Content-based focuses on the linguistic characteristics of textual reviews from users \cite{BAZZAZABKENAR2023120366} while behavior-based aims to extract the behavior of users' reviews according to the integration of users to others such as the number of reviews, product rating, the aspects, categories, and the properties of products \cite{SUN2024121236}. The combination of both approaches can exploit the behavior and the subjectivity of users in the product review, which is helpful for detecting whether the review is spam or not. There are several works proved that the integration of external information boosts the performance of the classification model such as the work in \cite{10.1145/2184436.2184437} integrates the prior knowledge of word sentiment to the LDA model, the work in \cite{10.1145/3394113} combines the product attributes with deep neural network for the sentiment analysis based on product reviews. In particular, Li \cite{10.1145/3453694} proposed a multimodel approach that combines not only the text but also the picture to extract the users' opinions from online tourism reviews, and the works in \cite{CHEN2023101326,ZHANG2023119454} constructed a graph neural network that treating the relationship between authors and the aspects of the review including the title, category, and the review content. These works are robust, but they require a complex representation of data to contribute the metadata of users' behavior with the reviews. In this paper, we proposed a simple method but efficient that can integrate the available external metadata features from products and users' reviews to enhance the ability of the spam detection system.

%% file: method.tex
\section{Methodology}
\label{method}

\subsection{The Dataset}

In \cite{vispamreviews}, the authors introduced the dataset and provided detailed descriptions of the labels for the spam detection task. \textbf{NO-SPAM} are regular reviews that provide helpful information for buyers to get an overview of the product. \textbf{SPAM} are reviews that are either not truthful or provide unhelpful information to users. These spam reviews are categorized into three types. The \textbf{SPAM-1 (fake review)} deceives users by providing negative comments about the product to damage the store's reputation or raising seeding to a product by hallucinating comments. Next, the \textbf{SPAM-2 (review on brand only)} only focuses on the brand, manufacturer, or seller except for the quality of the product. Finally, the \textbf{SPAM-3 (non-review)} reviews are which do not mention anything relevant to the products. The dataset in \cite{vispamreviews} comprises two tasks: \textbf{Task 1} for determining whether the reviews are spam or not spam, and \textbf{Task 2} for classifying the types of spam reviews. 

\begin{figure}[ht]
    \centering
    \resizebox{\textwidth}{!}{
    \begin{tikzpicture}
        \renewcommand\fbox{\fcolorbox{gray}{white}}
        \node[anchor=south west,inner sep=0] (image) at (0,0) {\fbox{\includegraphics[trim=0 0 0 0, clip, width=\textwidth, angle=0]{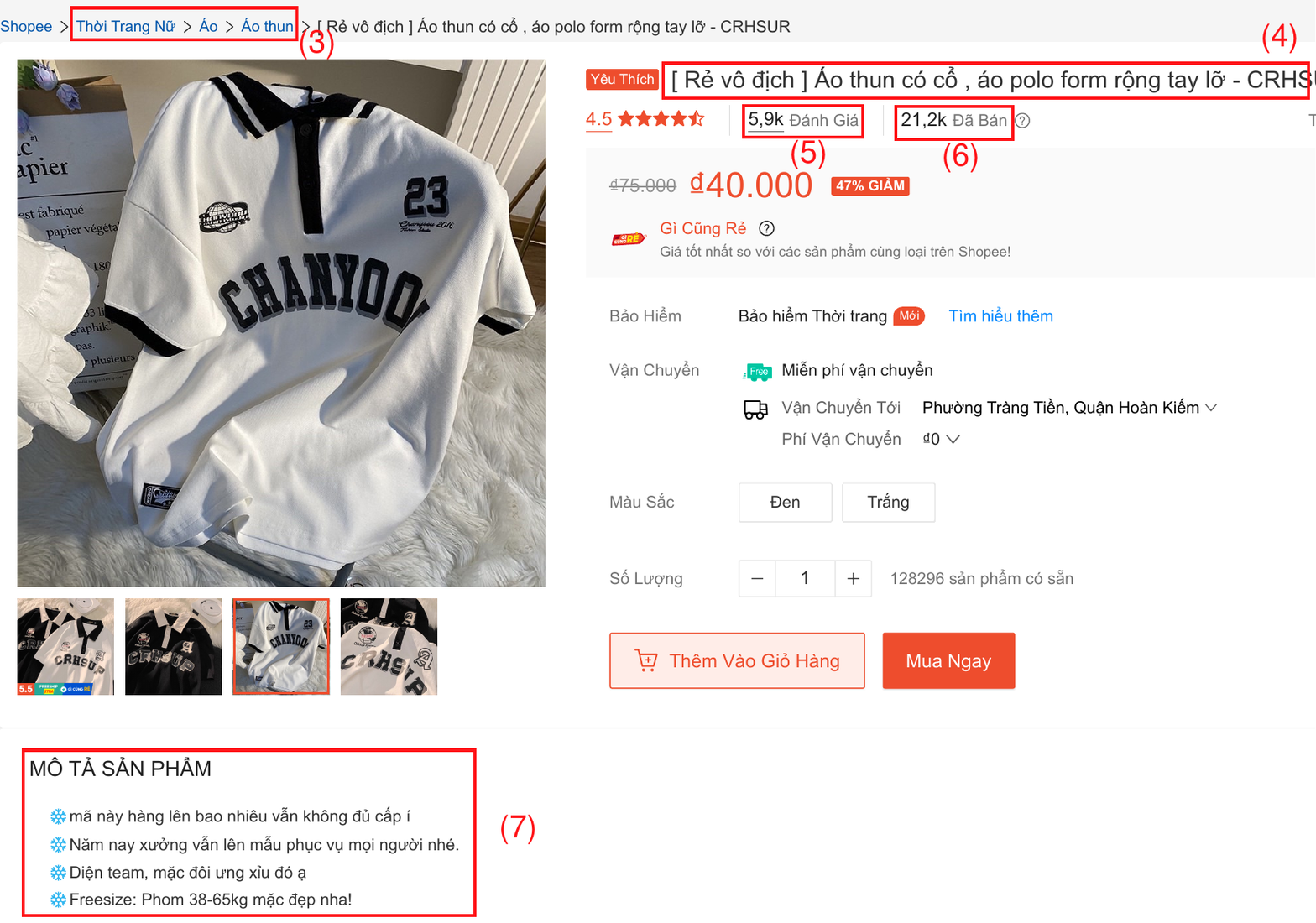}}};
        \begin{scope}[x={(image.south east)},y={(image.north west)}]
            

            \draw[red,thick,rounded corners] (0.06,0.935) rectangle (0.235,0.975);
            
            \node[draw=none] at (0.26, 0.92) {\textcolor{red}{(1)}};

            \draw[red,thick,rounded corners] (0.5,0.875) rectangle (0.995,0.925);
            
            \node[draw=none] at (0.9, 0.84) {\textcolor{red}{(2)}};

            \draw[red,thick,rounded corners] (0.56,0.835) rectangle (0.66,0.875);

            \node[draw=none] at (0.63, 0.81) {\textcolor{red}{(3)}};

            \draw[red,thick,rounded corners] (0.67,0.835) rectangle (0.765,0.875);

            \node[draw=none] at (0.73, 0.81) {\textcolor{red}{(4)}};

            \draw[red,thick,rounded corners] (0.035,0.015) rectangle (0.36,0.15);

            \node[draw=none] at (0.39, 0.08) {\textcolor{red}{(5)}};
            
        \end{scope}
        
    \end{tikzpicture}
    }
    \caption{The metadata is collected from Shopee.}
    \label{fig:shopee_metadata}
\end{figure}

In the ViSpamReviews v2 dataset, we have collected additional information related to the products (metadata) for each review without modifying the previous annotated label. Some of the information we have gathered from Shopee \footnote{\url{https://shopee.vn/}} is illustrated in Figure \ref{fig:shopee_metadata}, which encompasses product category (1), product name (2), number of ratings (3), number of sold items (4), and product description (5). In addition, we illustrate a review and its metadata in Table \ref{tab:example_review}. 

\begin{table}[ht!]
    \centering
    \renewcommand{\arraystretch}{1.25}
    \caption{An example review in the ViSpamReviews v2 dataset.}{
    \resizebox{\textwidth}{!}{
    \begin{tabular}{lp{14cm}}
        \toprule
        \textbf{Star rating} & 5 \\
        \midrule
        \textbf{Comment} & \textbf{Vietnamese:} Dùng tốt, sạc rất nhanh nên mau nóng máy, hạn chế ko dùng lúc đang sạc để khỏi phỏng tay. Sản phẩm hoàn thiện tốt về thiết kế, mình thấy chả có gì để chê. Lúc mình mua áp các loại mã giảm chỉ còn có 280k thui, freeship tikinow. Chỉ 1 giờ là giao tới nơi. 
        \newline (\textbf{English:} Performs well, charges very quickly but heats up quickly too, so it's recommended not to use while charging to avoid burning your hands. The product has a good design, I have nothing to complain about. I bought it with a discount code for only 280k VND and free shipping with Tikinow. It was delivered in just one hour.) \\
        \midrule
        \textbf{Category} & \textbf{Vietnamese:} Điện tử
        \newline (\textbf{English:} Electronics) \\
        \midrule
        \textbf{Product name} & \textbf{Vietnamese:} Adapter Sạc 1 Cổng Anker 18W/20W PowerPort III Nano Tích Hợp PowerIQ 3.0 - Hàng Chính Hãng
        \newline (\textbf{English:} Anker 18W/20W PowerPort III Nano 1-Port Charger Adapter with PowerIQ 3.0 Integration - Genuine Product) \\
        \midrule
        \textbf{Description} & \textbf{Vietnamese:} PowerPort III Nano Adapter Sạc 1 Cổng Anker 18W/20W PowerPort III Nano Tích Hợp PowerIQ 3.0 có thiết kế nhỏ gọn 18W tích hợp PowerIQ 3.0 mang đến tốc độ sạc nhanh chóng. Thiết kế nhỏ gọn tiện mang đi Với kiểu dáng nhỏ gọn chỉ 30g, bằng trọng lượng với pin tiểu AA; tuy nhỏ nhưng đây là loại adapter sạc mạnh mẽ mà bạn không ngờ đến. Nhờ vào việc thay đổi kích cỡ, adapter sạc với tốc độ nhanh dễ dàng để bỏ vào ví, túi khi ra ngoài, tiện lợi mang đi mọi nơi. Công nghệ sạc cao cấp Mang đến tốc độ sạc nhanh chóng cho mọi loại điện thoại và tablets có cổng kết nối USB-C (bao gồm cả iPhone và iPad) nhờ vào công nghệ Power Delivery. Tiện lợi cho mọi thiết bị...
        \newline (\textbf{English:} Anker PowerPort III Nano 1-Port 18W/20W Adapter with PowerIQ 3.0 Integration is designed with a compact size for fast charging. The small and lightweight design, weighing only 30g, which is equivalent to a AA battery, is a powerful charger that you would not expect. With a size change, the fast charging adapter is easy to put in your wallet or bag when you go out, and convenient to take anywhere. High-end charging technology ensures fast charging for all types of phones and tablets with a USB-C connection (including iPhones and iPads) thanks to Power Delivery technology. It is compatible with a wide range of devices...) \\
        \midrule
        \textbf{Number of sold} & 11873 \\
        \midrule
        \textbf{Number of reviews} & 3722 \\
        \midrule
        \textbf{Label} & NO-SPAM \\
        \midrule
        \textbf{Spam label} & -- \\
        \bottomrule
    \end{tabular}}}
    \label{tab:example_review}
\end{table}

\begin{table}[H]
    \centering
    \newcolumntype{P}[1]{>{\centering\arraybackslash}p{#1}}
    \caption{The number of labels for the reviews in each set after the resplit.}
    \begin{tabular}{c|P{1.5cm}|P{1.5cm}|P{1.5cm}}
     & \textbf{Train} & \textbf{Dev} & \textbf{Test}  \\
    \hline
     \textbf{NO-SPAM} & 10,490 & 1,182 & 2,911 \\
     \hline
     \textbf{SPAM-1} & 211 & 17 & 54 \\ 
     \hline
     \textbf{SPAM-2} & 1,103 & 120 & 309 \\ 
     \hline
     \textbf{SPAM-3} & 2,480 & 270 & 689 \\ 
     \hline
     \textbf{Total} & 14,284 & 1,589 & 3,963 \\ 
     \hline
    \end{tabular}
    \label{tab:label_distribution}
\end{table}

After collecting additional metadata, we resplit the dataset into train and test sets with a ratio of 8:2, respectively. In the training phase, we allocate 10\% of the training set to serve as the development set. The distribution of review samples for each label is depicted in Table \ref{tab:label_distribution}. Besides, some characteristics of the new dataset, such as the number of reviews, vocabulary size, and the average length of comments in each dataset, are presented in Table \ref{tab:vocabulary_size_set}.

\begin{table}[H]
    \renewcommand{\arraystretch}{1.25}
    \caption{The statistical characteristics of the new dataset (vocabulary size and average length are calculated based on word count).}
    \centering
    \newcolumntype{P}[1]{>{\centering\arraybackslash}p{#1}}
    \begin{tabular}{l|P{1.5cm}|P{1.5cm}|P{1.5cm}|P{1.5cm}}
         & \textbf{Train} & \textbf{Dev} & \textbf{Test} & \textbf{All} \\
         \hline
         \textbf{The number of reviews} & 14,284 & 1,589 & 3,963 & 19,836 \\
         \hline
         \textbf{Vocabulary size} & 17,196 & 4,414 & 8,144 & 21,405 \\
         \hline
         \textbf{Average length of comments} & 17.38 & 16.92 & 17.37 & 17.35 \\
         \hline
    \end{tabular}
    \label{tab:vocabulary_size_set}
\end{table}

\subsection{Metadata Analysis}

\begin{figure}[ht!]
    \definecolor{color1}{HTML}{2181fd}
    \definecolor{color2}{HTML}{00cc96}
    \definecolor{color3}{HTML}{fdc309}
    \definecolor{color4}{HTML}{fd3233}
    \centering
    \begin{tikzpicture}
    \begin{axis}[
        ybar stacked,
        width=0.8*\textwidth,
        height=9cm,
        xlabel={},
        bar width=15pt,
        ylabel={Number of reviews},
        ylabel style={font=\small},
        symbolic x coords={Fashion,Electronic,Others,Phone - Computer,Beauty,Home - Life,Toy,Sport,Camera,Vehicel,Footwear,Book},
        xticklabel style={rotate=90, anchor=near xticklabel, font=\small},
        yticklabel style={font=\small},
        xtick=data,
        legend cell align=center,
        legend style={
                draw=none,
                fill=none,
                align=center,
                at={(0.5,1.10)},
                anchor=north,
                column sep=1ex,
                legend columns=-1,
                font=\fontsize{8}{6}\selectfont
        },
        legend image code/.code={\draw [#1] (0cm,-0.1cm) rectangle (0.5cm,0.25cm);},
        every node near coord/.append style={font=\small}
    ]
    \addplot[style={color1, fill=color1, mark=none}]
        coordinates {(Fashion, 1947) (Electronic, 2476) (Others, 1662) (Phone - Computer, 1752) (Beauty, 1323) (Home - Life, 1354) (Toy, 1212) (Sport, 868) (Camera, 647) (Vehicel, 565) (Footwear, 389) (Book, 405)};

    \addplot[style={color2, fill=color2, mark=none}]
        coordinates {(Fashion, 65) (Electronic, 8) (Others, 61) (Phone - Computer, 19) (Beauty, 24) (Home - Life, 39) (Toy, 15) (Sport, 9) (Camera, 7) (Vehicel, 8) (Footwear, 14) (Book, 13)};

    \addplot[style={color3, fill=color3, mark=none}]
        coordinates {(Fashion, 99) (Electronic, 199) (Others, 304) (Phone - Computer, 170) (Beauty, 229) (Home - Life, 107) (Toy, 192) (Sport, 32) (Camera, 68) (Vehicel, 73) (Footwear, 34) (Book, 25)};

    \addplot[style={color4, fill=color4, mark=none}]
        coordinates {(Fashion, 853) (Electronic, 222) (Others, 435) (Phone - Computer, 366) (Beauty, 315) (Home - Life, 338) (Toy, 256) (Sport, 176) (Camera, 83) (Vehicel, 125) (Footwear, 137) (Book, 116)};
    
    \legend{NO-SPAM, SPAM-1, SPAM-2, SPAM-3}
    \end{axis}
    \end{tikzpicture}
    \caption{Distribution of labels for reviews by product category.}
    \label{fig:spamlabel_vs_categories}
\end{figure}
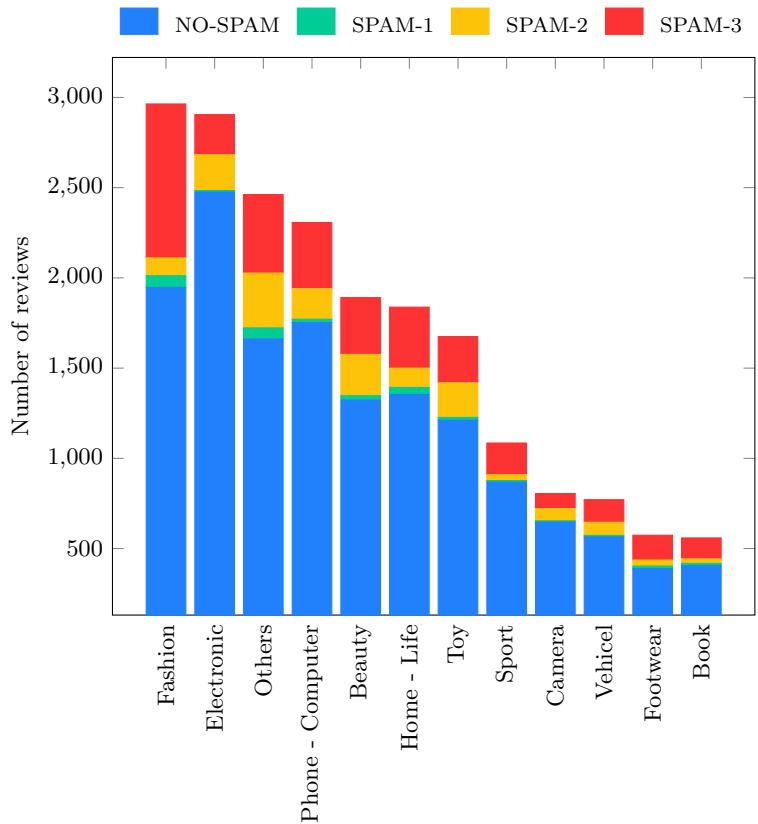

The gathered attributes encompass the product category, product name, product description, number of ratings, and number of sold. E-commerce platforms have different strategies to categorize various product categories, but there are still commonalities among them. We built a dictionary to merge the categories on Shopee and Tiki together, and then we analyzed the number of spam reviews for each product category in Table \ref{tab:label_vs_categories}. Moreover, Figure \ref{fig:spamlabel_vs_categories} depicts a graphical representation of the label ratios for classifying different types of spam reviews.

\begin{table}[H]
    \centering
    \renewcommand{\arraystretch}{1.25}
    \caption{Statistics of the number of labels by each product category.}{
    \begin{tabular}{lcccc}
        \toprule
        \textbf{Category} & \textbf{Total} & \textbf{NO-SPAM} & \textbf{SPAM} & \textbf{Spam rate (\%)} \\
        \midrule
        Fashion & 2,964 & 1,947 & 1,017 & 34.31 \\
        Electronic & 2,905 & 2,476 & 429 & 14.77  \\
        Others & 2,462 & 1,662 & 800 & 32.49  \\
        Phone - Computer & 2,307 & 1,752 & 555 & 24.06   \\
        Beauty & 1,891 & 1,323 & 568 & 30.04  \\
        Home - Life & 1,838 & 1,354 & 484 & 26.33  \\
        Toy & 1,675 & 1,212 & 463 & 27.64  \\
        Sport & 1,085 & 868 & 217 & 20.00  \\
        Camera & 805 & 647 & 158 & 19.63  \\
        Vehicel & 771 & 565 & 206 & 26.72  \\
        Footwear & 574 & 389 & 185 & 32.23  \\
        Book & 559 & 405 & 154 & 27.55  \\
        \bottomrule
    \end{tabular}}
    \label{tab:label_vs_categories}
\end{table}

According to Table \ref{tab:label_vs_categories}, there are a total of 12 categories after the consolidation. It is observed that the occurrence of spam reviews is not evenly distributed across each category, with Fashion, Beauty, Footwear, and Others categories exhibiting higher spam ratios compared to the rest. This is reasonable as there are some highly popular products with a large volume of sales, resulting in a higher proportion of spam reviews. Additionally, the distribution of spam labels is also disparate, with some categories having a higher prevalence of SPAM-2 and SPAM-3 labels, such as Fashion, Beauty, Toy, and Others. Hence, this information proves useful in both spam review classification and identifying types of spam review tasks.

In the task of identifying the type of spam reviews, prediction models often encounter higher mispredictions on the SPAM-2 label due to incomplete information provided in the comments of reviews. Thus, it is necessary to consider which attributes are important to combine between the product name and product description. The average length of textual attributes for each label of reviews, computed at the word level, is presented in Table \ref{tab:label_distribution_spam_type}.

\begin{table}[H]
    \renewcommand{\arraystretch}{1.25}
    \newcolumntype{P}[1]{>{\centering\arraybackslash}p{#1}}
    \caption{The average length of textual attributes on the train set (calculated based on word count).}{
    \resizebox{\textwidth}{!}{
    \begin{tabular}{l|P{2cm}|P{1.6cm}|P{1.6cm}|P{1.6cm}}
     & \textbf{NO-SPAM} & \textbf{SPAM-1} & \textbf{SPAM-2} & \textbf{SPAM-3}  \\
    \hline
     \textbf{Average length of comments} & 17.74 & 25.12 & 13.82 & 16.76 \\
     \hline
     \textbf{Average length of product names} & 15.11 & 14.36 & 15.07 & 15.06 \\ 
     \hline
     \textbf{Average length of descriptions} & 239.30 & 235.71 & 248.76 & 225.42 \\ 
     \hline
    \end{tabular}
    }}
    \label{tab:label_distribution_spam_type}
\end{table}

\begin{figure}[htb]
    \centering
    \resizebox{\textwidth}{!}{
    \centering
    \begin{minipage}[t]{.45\textwidth}
        \includegraphics[scale=0.4]{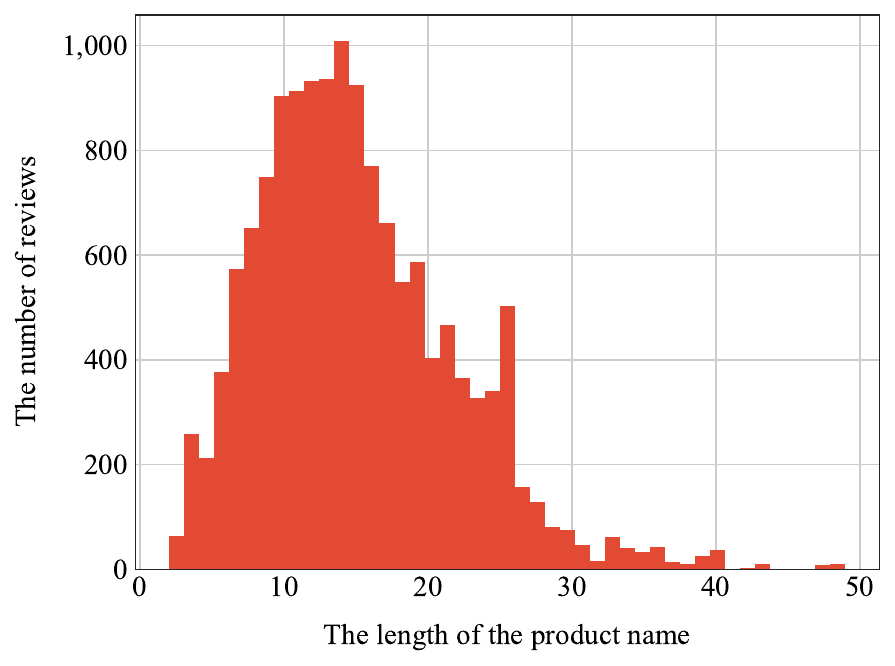}
    \end{minipage}
    \hspace{.6cm}
    \begin{minipage}[t]{.45\textwidth}
        \includegraphics[scale=0.4]{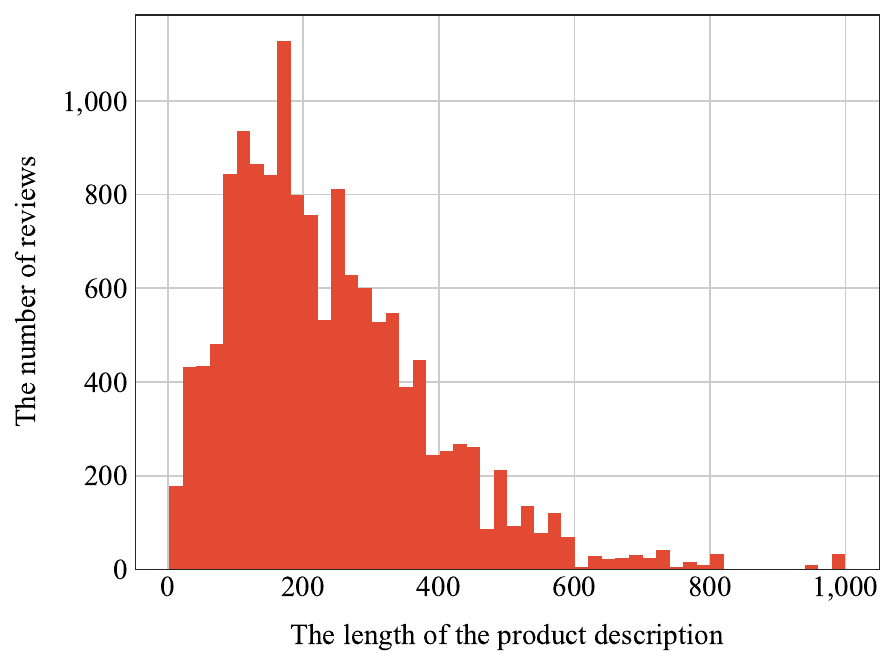}
    \end{minipage}
    }
    \caption{Distribution of product name and product description lengths (calculated based on word count) on the training set.}
    
    \label{fig:distribution_length_product_name_description}
\end{figure}

According to Table \ref{tab:label_distribution_spam_type}, we need to pay attention to the length of the SPAM-2 label, which has the shortest comment length of only 13.82, while the product description has a longer length compared to the product name, with average lengths of 15.07 and 248.76, respectively. Additionally, we also illustrate the length distribution of product names and product descriptions on the training set in Figure \ref{fig:distribution_length_product_name_description}. Comparing product names and product descriptions, we observe a significant difference, where the length of product descriptions ranges from 100 to 400, while the length of product names only ranges from 5 to 25. Specifically, the product name usually has a shorter length, typically consisting of brand names and product codes, whereas the product description is often written by sellers to provide detailed information about the characteristics, features, and uses of the product, resulting in a longer average length. Comments of reviews typically have a stronger association with product descriptions rather than product names (which primarily relate to brand names in the SPAM-2 labels). Therefore, we employ the product description as a supplementary text attribute to integrate into the classification model.

We consider the two remaining attributes: number of sold and number of ratings. Due to differing market shares, the dissimilarities in the number of buyers across the two platforms, result in variations in the quantities of sales and reviews. Consequently, these attributes are not suitable for integration into the model if the reviews are not from the same e-commerce platform.

\subsection{Metadata Integration}
\begin{figure}[h]
    \centering
    \includegraphics[width=\textwidth]{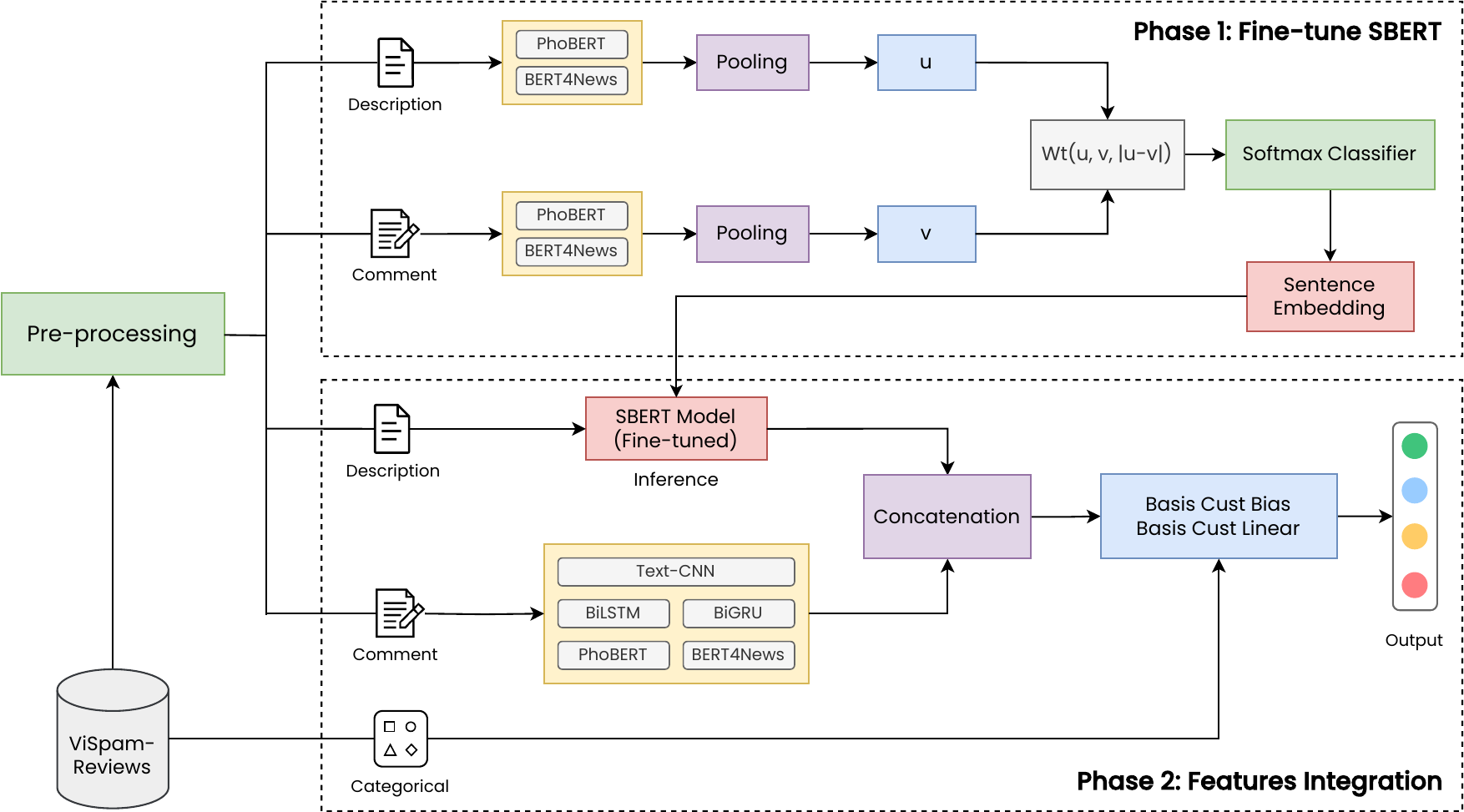}
    \caption{An overview of the proposed method for integrating metadata in spam reviews detection.}
    \label{fig:proposal_method}
\end{figure}

We propose a method to integrate metadata for the task of classifying spam reviews, as shown in Figure \ref{fig:proposal_method}. In this approach, we leverage metadata consisting of textual attributes and categorical variables, to enhance the classification of spam reviews\footnote{Source code is available at \url{https://github.com/sonlam1102/vispamdetection}}.

To integrate metadata into the spam review classification model, we present two distinct but interconnected phases as follows:
\begin{itemize}
    \item \textbf{Phase 1 - Fine-tune SBERT:} Firstly, for the textual attribute, we fine-tune an SBERT (Sentence-BERT) model \cite{reimers-2019-sentence-bert} to learn representations capturing the relationship between the product description and user comments. SBERT is a deep learning model pretrained to generate semantically meaningful sentence embeddings. Fine-tuning the model with labeled reviews enhances its ability to comprehend the semantic context of product descriptions, aiding in the determination of their correlation with user comments. This training is conducted only once, after which the fine-tuned model can be used to generate features for subsequent training stages and during inference on real data, resulting in savings in terms of cost and training time.

    \item \textbf{Phase 2 - Features Integration:} This is the stage where we integrate metadata features to train the label classification model for reviews. In this stage, we simply concatenate the embedded vector of the product description which is inferred from the SBERT model with the textual feature, which in this case is the user comment. We create a fused representation of the textual information by concatenating these two vectors. Subsequently, we incorporate the product category into the previously created representation. This is achieved using a proposed method by Kim et al \cite{kim-etal-2019-categorical}, we experimented with two methods of combining categorical features for the highest effectiveness on the Yelp 2013 dataset \cite{tang-etal-2015-learning} as presented by the author, namely at the bias vector level and linear transformation. These two levels of combination correspond to the Basis Cust Bias and Basis Cust Linear block in our illustration. Further details on this method will be presented below. Finally, after obtaining the representation of the textual information and metadata, we employ a softmax classifier to predict the label.
\end{itemize}

Note that fine-tuning the SBERT model is solely for generating features for product description texts accompanying each review comment, and it is reused to produce features for product descriptions for later inference, combined with the classifier when predicting labels for a review. Both the SBERT model and the classifier model are trained on labeled data and provide the final decision on the label of the review based on this fused representation. A significant highlight of the method presented above is that it helps save computational costs and training time, alongside its flexibility to combine with various other features including text and categorical features.

\subsubsection{Product Description}

In the analysis errors of the previous publication \cite{vispamreviews}, we indicated that relying solely on the user's comment is inadequate for the classifier to predict the label of the reviews accurately. We believe that the product description is the most suitable attribute to indicate whether the user's comment is related to the product or not. As mentioned in the label descriptions, a review is considered non-spam if the user's comment is helpful and relevant to the product. In contrast, spam reviews typically have content that is unrelated to the product, especially in the case of spam type 3. Hence, by representing the product description as a vector, it becomes possible to capture the similarity between it and the comment, allowing for their combination into the classifier and potentially improving the model's accuracy.

There are many methods that have been researched for sentence embedding, and SBERT is a model proposed to utilize BERTology models (BERT / RoBERTa) \cite{rogers-etal-2020-primer} for generating document representations. This approach has been demonstrated to provide the best effectiveness in learning the semantic relationships between documents. Taking advantage of the benefits of this model, we performed fine-tuning on SBERT using two pre-trained Vietnamese models comprised of PhoBERT \cite{nguyen-tuan-nguyen-2020-phobert} and BERT4News \cite{nguyen2021nlpbk}, in order to generate embedding vectors for product descriptions.

Initially, the comment and the corresponding product description are fed into BERTology models, an operation added to SBERT compared to regular BERT, which includes pooling to derive a fixed-sized sentence embedding. We employ a pooling strategy MEAN pooling to compute the mean of all output vectors, thereby generating two vectors of equal size: the comment embedding vector ($u$) and the product description embedding vector ($v$). Subsequently, these sentence embeddings are concatenated with the element-wise difference $|u-v|$ and multiply it with the trainable weight $W_t \in \mathbb{R}^{3n \times k}$, finally passed into a softmax classifier as in the equation:

\begin{equation*}
    \label{eq:sbert}
    o = softmax(W_t(u, v, |u-v|))
\end{equation*}

With $n$ is the dimension of the sentence embeddings and $k$ is the number of labels. The loss function used is Cross-Entropy.

\subsubsection{Product Category}

Customized text classification is introduced to customize the classifier based on categorical metadata information rather than only using text features. We incorporate a product category of the reviews, which is a common attribute found in any e-commerce platform. We observe that some spam reviews have a higher frequency of occurrence in certain product categories. Therefore, we combine this information to enhance the classification performance of the models. There are multiple methods to integrate the categorical variables into the classification model at different levels, including customizing the bias vector, customizing the word embeddings \cite{tang-etal-2015-learning}, customizing the attention pooling \cite{attention_pooling2016}, customizing on the linear transformation and customize on the encoder weights \cite{kim-etal-2019-categorical}. The method of customizing on the linear transformation has demonstrated effectiveness on some datasets. Therefore, we experiment with this method on our ViSpamReviews v2 dataset and we also evaluate the effectiveness of customizing on the bias vector.

In general, in the classifier, we have a text representation vector $d$ and category vectors $c_1, c_2,...,c_m$, a learned weight matrix $W^{(c)}$ and a bias vector $b^{(c)}$. For customizing on bias vector, instead of using a single bias vector $b^{(c)}$ in the logistic regression classifier, the model uses additional multiple bias vectors for each category. This method is similar to concatenating the features of the categorical variable with the text vector $d$ and the derivative is given by the following equation:

\begin{equation}
    y'=W^{(c)}[d;c_1;c_2;...;c_m]+b^{(c)}
    \label{eq:derivative_bias_vector}
\end{equation}

As for customizing on linear transformation, instead of using a single weight matrix $W^{(c)}$ in the logistic regression classifier, the model uses different weight matrices for each category, and the derivative is given by the equation:

\begin{equation}
    y'=W_{c_1}^{(c)}d+W_{c_2}^{(c)}d+...+W_{c_m}^{(c)}d+b^{(c)}
    \label{eq:derivative_linear_transformation}
\end{equation}

In addition to the conventional customized methods, basis-customized is also proposed to address some of its limitations. This method use a trainable set of bias vectors $B=\{b_1,b_2,...,b_d\}$, with $d << dim$ ($dim$ is the dimension of the original weights). The vector search space $V$ that contains all the optimal customized weights $v_c$, such that $B$ is the basis of $V_c$. The vectors $v \in V$ are calculated using the equation:

\begin{equation}
    v_c=\sum_{}\mathop{}_{i}\gamma_i * b_i
\end{equation}

Where $\gamma$ represents the coefficients and are calculated based on the following equations:

\begin{equation}
    \begin{gathered}
    z_i=q^{T}k_i\\
    \gamma_i = \frac{exp(z_i)}{\sum_{i} exp(z_j)}
    \end{gathered}
\end{equation}

With $q$ is the query vector that concatenated category vectors $q=[c_1;c_2;...;c_m]$ and $k_i \in K | K=\{k_1,k_2,...,k_d\}$ is a trainable set of key vectors.

%% file: results.tex
\section{Results and Discussion}
\label{result}

\subsection{Data Preprocessing}

\begin{figure}[htbp]
    \centering
    \includegraphics[trim=0 0 0 0, clip, width=0.8\textwidth, angle=0]{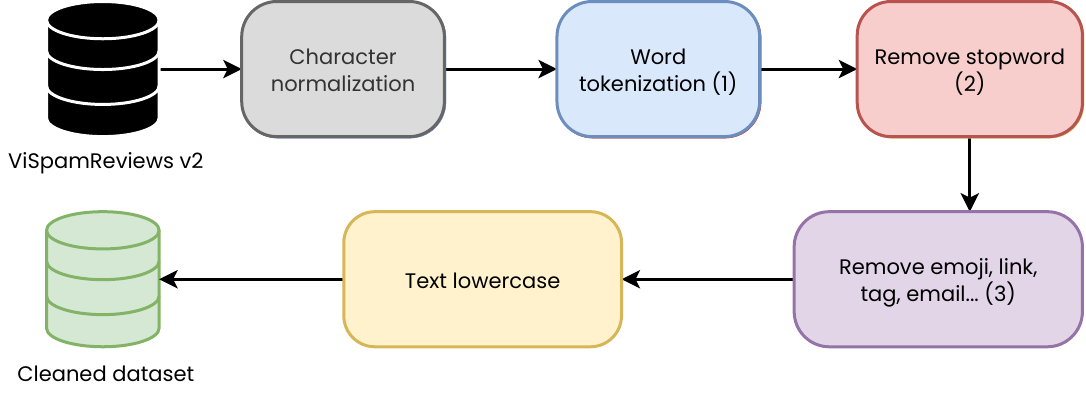}
    \caption{Data preprocessing pipeline.}
    \label{fig:data_preprocessing_pipeline}
\end{figure}

The reviews in the dataset are created by users to be posted on an online e-commerce platform, which requires preprocessing before training the model. We perform data cleaning of the comment sentences and product descriptions, following the process illustrated in Figure \ref{fig:data_preprocessing_pipeline}.

The steps in the process are described as follows:

\begin{itemize}
    \item \textbf{Step 1:} Text normalization to a uniform format. We observed that some comments, despite having similar structural content, are not considered identical, potentially due to the presence of different Unicode characters. Some tokenization tools may struggle to accurately recognize and tokenize this text. As a result, we standardize the user comments and product description content to ensure uniformity.

    \item \textbf{Step 2:} Word tokenization using the VnCoreNLP tool \footnote{\url{https://github.com/vncorenlp/VnCoreNLP}}. Pretrained models can be trained with tokenized text, so we preprocess the data to ensure compatibility with the model. Additionally, in some cases, word tokenization can enhance the performance of models.

    \item \textbf{Step 3:} Removing Vietnamese stopwords\footnote{\url{https://github.com/stopwords/vietnamese-stopwords}}. These are frequently encountered words in the text that hold little semantic value. By eliminating stopwords, the model can better learn from the meaningful content within the text. Therefore, we remove Vietnamese stopwords on both the comment and description attributes.

    \item \textbf{Step 4:} Removing emojis, links, tags, emails... This step is specifically applied to the product description. In the case of user comments, emojis are also one of the factors that affect the label of the review, and comments that contain only emails or tags, or links are classified as spam according to the annotation guidelines. Hence, this step is exclusively conducted on the product description, where these types of content are commonly found.

    \item \textbf{Step 5:} Converting all characters in the text to lowercase.
\end{itemize}

\subsection{Experiment Settings}

We conducted experiments on two tasks as outlined in the previous publication \cite{vispamreviews}: determining whether the reviews are spam or not (\textbf{Task 1}) and identifying the type of spam reviews (\textbf{Task 2}). In order to evaluate the efficacy of the metadata integration methods, we carried out experiments on different deep neural network (DNN) models, specifically TextCNN \cite{kim-2014-convolutional}, BiLSTM \cite{hochreiter1997long} and BiGRU \cite{cho-etal-2014-properties}. In addition, we also employed transfer learning models in our experiments, namely BERTology models with two pre-trained consisting of PhoBERT and BERT4News.

In the fine-tuning phase of the SBERT model (Phase 1), we set the max length sequence is 100 for both product descriptions and comments. Then, we fine-tuned SPhoBert and SBert4News for 20 epochs and mini-batch size equal to 16. Following that, we embedded the product descriptions and concatenated them with the encoder of the baseline models. As for the product category using the basis-customized method, we set 64 dimensions for the category vector.

In the experimentation phase of training the classifier with metadata integration (Phase 2), we also set the max length sequence of comments and descriptions to 100, the seed is set equal to 42 for all experiments. For DNN models, we trained them for 40 epochs with mini-batch size of 256. We employ the Early Stopping technique with patience is 10 and saved the best-performing model. As for the BERT models, we performed fine-tuning \textbf{PhoBERT} and \textbf{BERT4News} for 10 epochs, batch size equal to 16, and employed Early Stopping with patience is 3. The Cross-Entropy loss function was used. The DNN models were configured with the following specific parameters:

\begin{itemize}
    \item \textbf{TextCNN}: We utilized three Conv2D layers with 32 filters of sizes $[2, 3, 5]$, the activation function is softmax, and the dropout is set to 0.5 to prevent overfitting.

    \item \textbf{BiLSTM} and \textbf{BiGRU}: We employed the Bidirectional layer in both the LSTM and GRU encoder with the hidden size of 64. We then concatenated average pooling and max pooling, used the ReLU activation function, and also set the dropout to 0.5 as TextCNN.
\end{itemize}

Our experiments are performed on the Kaggle platform \footnote{\url{https://www.kaggle.com/}} with hardware specifications are Intel(R) Xeon(R)CPU @ 2.30GHz and NVIDIA Tesla P100 GPU with 16GB RAM.

\subsection{Evaluation Metrics}
\label{metrics}
To assess the performance of the models, we use Accuracy and F1-score, which are two essential metrics for classification problems \cite{Sokolova2009ASA}. These metrics serve to gauge the model's precision in classifying data points and also provide insights into the classification performance for each class. The following section presents the methodologies applied to compute these metrics in the context of each specific task.

\subsubsection{Task 1}
In this task, our objective is to classify reviews to determine whether they are spam or not. Therefore, this is a binary classification problem with two classes: spam and non-spam. The metrics are calculated using the formulas provided below.

\begin{equation}
    \text{Accuracy} = \frac{tp + tn}{tp + tn + fp + fn}
\end{equation}

\begin{equation}
    \text{Precision} = \frac{tp}{tp + fp}
\end{equation}

\begin{equation}
    \text{Recall} = \frac{tp}{tp + fn}
\end{equation}

\begin{equation}
    \text{F1-score} = \frac{2 \times \text{Precision} \times \text{Recall}}{\text{Precision} + \text{Recall}}
\end{equation}

Where $tp$ (true positive) is the number of reviews predicted to match the non-spam label, $fp$ (false positive) is the number of reviews with the spam label predicted as non-spam, $tn$ (true negative) is the number of reviews predicted to match the spam label, $fn$ (false negative) is the number of reviews with the non-spam label predicted as spam, respectively.
Specifically, the F1-score is a combination of the precision and recall metrics, aiming to balance the model's performance in cases where the data is imbalanced.

\subsubsection{Task 2}

As for this task, our goal is to determine the type of spam reviews. there are three spam labels and one non-spam label, and the total number of labels to be classified is four. Hence, this task is a multi-class classification problem. Similar to the binary classification problem, Accuracy and F1-score are also adjusted to suit the assessment objectives for multi-class classification. These metrics are calculated according to the formulas presented below.

\begin{equation}
    \text{Accuracy} = \frac{\text{Number of correct predictions}}{\text{Total number of predictions}}
\end{equation}

\begin{equation}
    \text{Precision} = \frac{1}{n} \sum_{i=1}^{n} \frac{tp_i}{tp_i + fp_i}
\end{equation}

\begin{equation}
    \text{Recall} = \frac{1}{n} \sum_{i=1}^{n} \frac{tp_i}{tp_i + fn_i}
\end{equation}

\begin{equation}
    \text{F1-score} = \frac{2 \times \text{Precision} \times \text{Recall}}{\text{Precision} + \text{Recall}}
\end{equation}

Where n is the number of classes in the classification problem. In this task, n equals 4. $tp_i$ represents the true positive for the i-th class, $fp_i$ is the false positive, $tn_i$ is the true negative and $fn_i$ is the false negative, respectively. In the formulas, the values of precision and recall metrics are calculated for each class, and then the averages are computed. Finally, the F1-score is determined based on the resulting precision and recall values.

\subsection{Empirical Results}
We use two metrics, namely Accuracy and macro-averaged F1-score as presented in section \ref{metrics}, to evaluate the models. Table \ref{tab:experimental_results} illustrates the experimental results of the baseline models on the ViSpamReviews v2 corpus and their comparison with models incorporating metadata. According to Table \ref{tab:experimental_results}, among the baseline models (without metadata integration), PhoBERT achieves the highest Accuracy and macro F1-score, obtaining 89.23\% and 85.58\% respectively for Task 1, and the corresponding results are 87.79\% and 71.56\% for Task 2. Considering only DNN models, BiLSTM demonstrates the highest Accuracy and F1-macro, reaching 76.55\% and 61.04\%, respectively. These results indicate that transfer learning models outperform conventional deep learning models in terms of performance.

Initially, we only incorporated the product category into the classifier using customized-text classification methods, including customizing the bias vector (bias cust and bias basis cust), and customizing the linear transformation matrix (linear cust and linear basis cust). The DNN models showed a significant increase in accuracy when incorporating the category. Specifically, TextCNN demonstrated a substantial improvement by achieving a 2.69\% and 5.32\% increase when combined with linear basis cust. On the other hand, BiLSTM and BiGRU don't show considerable improvements on Task 2 and even performance decreases in some cases. Nevertheless, these models demonstrated significant accuracy enhancement on Task 1, BiGRU was combined with linear basis cust, and the accuracy improved considerably, with respective increases of 2.89\% and 3.38\% measured by F1-macro, corresponding to Task 1 and Task 2. In contrast, the BERT models experienced a decrease in accuracy when incorporating the category. Although the decrease was not substantial and also there is a slight increase was observed, it is evident that the category is effective only when combined with DNN models, while it is not effective with BERT models.

\begin{table}[ht]
    \centering
    \renewcommand{\arraystretch}{1.2}
    \caption{The experimental results on ViSpamReviews v2 dataset.}{
    \resizebox{.9\textwidth}{!}{
    \begin{tabular}{l|l|l|cc|cc}
        \hline
        \multirow{2}{*}{\textbf{Model}} & \multicolumn{2}{c|}{\multirow{2}{*}{\textbf{Metadata}}} & \multicolumn{2}{c|}{\textbf{Accuracy (\%)}} & \multicolumn{2}{c}{\textbf{F1-macro (\%)} } \\ 
        
        \cline{4-7}
        & \multicolumn{2}{c|}{} & \textbf{Task 1} & \textbf{Task 2} & \textbf{Task 1} & \textbf{Task 2} \\ \hline
        
        \multirow{9}{*}{TextCNN} & \multicolumn{2}{l|}{Original} & 82.26 & 78.58 & 77.18 & 60.64  \\
        \cline{2-7}
        
        & \multirow{4}{*}{Category} & + bias cust & \textbf{85.24($\uparrow\textbf{2.98}$)} & \textbf{80.55($\uparrow\textbf{1.97}$)} & \textbf{79.93($\uparrow\textbf{2.75}$)} & \textbf{64.28($\uparrow\textbf{3.64}$)}  \\
        
        & & + bias basis cust & \textbf{84.36($\uparrow\textbf{2.10}$)} & \textbf{80.47($\uparrow\textbf{1.89}$)} & \textbf{79.64($\uparrow\textbf{2.46}$)} & \textbf{63.77($\uparrow\textbf{3.13}$)} \\
        
        & & + linear cust & 81.86($\downarrow0.40$) & \textbf{79.33($\uparrow\textbf{0.75}$)} & 75.84($\downarrow1.34$) & 59.22($\downarrow1.42$) \\
        
        & & + linear basis cust & \textbf{84.38($\uparrow\textbf{2.12}$)} & \textbf{82.44($\uparrow\textbf{3.86}$)} & \textbf{79.87($\uparrow\textbf{2.69}$)} & \textbf{65.96($\uparrow\textbf{5.32}$)} \\
        
        \cline{2-7}
        
        & \multirow{2}{*}{Description} & + SPhoBert & \textbf{83.02($\uparrow\textbf{0.76}$)} & \textbf{79.84($\uparrow\textbf{1.26}$)} & \textbf{78.48($\uparrow\textbf{1.30}$)} & \textbf{60.84($\uparrow\textbf{0.20}$)} \\
        
        & & + SBert4News & \textbf{83.35($\uparrow\textbf{1.09}$)} & \textbf{79.54($\uparrow\textbf{0.96}$)} & \textbf{78.80($\uparrow\textbf{1.62}$)} & \textbf{62.67($\uparrow\textbf{2.03}$)} \\

        \cline{2-7}
        & \multirow{2}{*}{Both} & + linear basis cust & \multirow{2}{*}{\textbf{83.85($\uparrow\textbf{1.59}$)}} & \multirow{2}{*}{\textbf{80.32($\uparrow\textbf{1.74}$)}} & \multirow{2}{*}{\textbf{78.73($\uparrow\textbf{1.55}$)}} & \multirow{2}{*}{\textbf{64.97($\uparrow\textbf{4.33}$)}} \\
        
        & & + SBert4News &  & & & \\
        
        \hline
        
        \multirow{9}{*}{BiLSTM} & \multicolumn{2}{l|}{Original} & 82.94 & 78.37 & 76.55 & 61.04 \\

        \cline{2-7}

        & \multirow{4}{*}{Category} & + bias cust & \textbf{84.13($\uparrow\textbf{1.19}$)} & \textbf{79.59($\uparrow\textbf{1.22}$)} & \textbf{78.72($\uparrow\textbf{2.17}$)} & \textbf{61.11($\uparrow\textbf{0.07}$)} \\
        
        & & + bias basis cust & \textbf{84.13($\uparrow\textbf{1.19}$)} & 78.20($\downarrow0.17$) & \textbf{77.85($\uparrow\textbf{1.30}$)} & \textbf{61.53($\uparrow\textbf{0.49}$)} \\
        
        & & + linear cust & \textbf{83.83($\uparrow\textbf{0.89}$)} & \textbf{79.99($\uparrow\textbf{1.62}$)} & \textbf{79.43($\uparrow\textbf{2.88}$)} & \textbf{61.10($\uparrow\textbf{0.06}$)} \\
        
        & & + linear basis cust & \textbf{85.21($\uparrow\textbf{2.27}$)} & \textbf{79.74($\uparrow\textbf{1.37}$)} & \textbf{79.93($\uparrow\textbf{3.38}$)} & 61.01($\downarrow0.03$) \\
        
        \cline{2-7}
        
        & \multirow{2}{*}{Description} & + SPhoBert & \textbf{84.13($\uparrow\textbf{1.19}$)} & 77.16($\downarrow1.21$) & \textbf{79.33($\uparrow\textbf{2.78}$)} & \textbf{61.13($\uparrow\textbf{0.09}$)} \\
        
        & & + SBert4News & \textbf{84.33($\uparrow\textbf{1.39}$)} & \textbf{80.72($\uparrow\textbf{2.35}$)} & \textbf{79.44($\uparrow\textbf{2.98}$)} & \textbf{65.77($\uparrow\textbf{4.73}$)} \\

        \cline{2-7}
        & \multirow{2}{*}{Both} & + linear basis cust & \multirow{2}{*}{\textbf{84.51($\uparrow\textbf{1.57}$)}} & \multirow{2}{*}{\textbf{80.82($\uparrow\textbf{2.45}$)}} & \multirow{2}{*}{\textbf{80.43($\uparrow\textbf{3.88}$)}} & \multirow{2}{*}{\textbf{63.64($\uparrow\textbf{2.60}$)}} \\
        
        & & + SBert4News &  & & & \\
        
        \hline
        
        \multirow{9}{*}{BiGRU} & \multicolumn{2}{l|}{Original} & 83.12 & 77.44 & 77.86 & 60.95 \\
        \cline{2-7}

        & \multirow{4}{*}{Category} & + bias cust & \textbf{84.28($\uparrow\textbf{1.16}$)} & 77.16($\downarrow0.28$) & \textbf{79.94($\uparrow\textbf{2.08}$)} & 57.29($\downarrow3.66$) \\
        
        & & + bias basis cust & \textbf{84.58($\uparrow\textbf{1.46}$)} & \textbf{77.77($\uparrow\textbf{0.33}$)} & \textbf{79.71($\uparrow\textbf{1.85}$)} & 59.50($\downarrow1.45$) \\
        
        & & + linear cust & 82.64($\downarrow0.48$) & 77.24($\downarrow0.20$) & 77.10($\downarrow0.76$) & 58.22($\downarrow2.73$) \\
        
        & & + linear basis cust & \textbf{84.83($\uparrow\textbf{1.71}$)} & \textbf{79.81($\uparrow\textbf{2.37}$)} & \textbf{80.75($\uparrow\textbf{2.89}$)} & \textbf{64.33($\uparrow\textbf{3.38}$)} \\
        
        \cline{2-7}
        
        & \multirow{2}{*}{Description} & + SPhoBert & \textbf{84.05($\uparrow\textbf{0.93}$)} & \textbf{79.61($\uparrow\textbf{2.17}$)} & \textbf{79.53($\uparrow\textbf{1.67}$)} & \textbf{62.39($\uparrow\textbf{1.44}$)} \\
        
        & & + SBert4News & \textbf{83.70($\uparrow\textbf{0.58}$)} & 77.01($\downarrow0.43$) & \textbf{79.34($\uparrow\textbf{1.48}$)} & \textbf{62.39($\uparrow\textbf{1.44}$)} \\

        \cline{2-7}
        & \multirow{2}{*}{Both} & + linear basis cust & \multirow{2}{*}{\textbf{84.89($\uparrow\textbf{1.77}$)}} & \multirow{2}{*}{\textbf{81.30($\uparrow\textbf{3.86}$)}} & \multirow{2}{*}{\textbf{79.71($\uparrow\textbf{1.85}$)}} & \multirow{2}{*}{\textbf{63.37($\uparrow\textbf{2.42}$)}} \\
        
        & & + SPhoBert &  & & & \\
        
        \hline
        
        \multirow{8}{*}{PhoBERT} & \multicolumn{2}{l|}{Original} & 89.23 & 87.79 & 85.58 & 71.56 \\

        \cline{2-7}

         & \multirow{4}{*}{Category} & + bias cust & \textbf{89.35($\uparrow\textbf{0.12}$)} & \textbf{88.54($\uparrow\textbf{0.75}$)} & \textbf{86.32($\uparrow\textbf{0.74}$)} & 71.07($\downarrow0.49$) \\
        
        & & + bias basis cust & 88.70($\downarrow0.53$) & 87.61($\downarrow0.18$) & \textbf{85.65($\uparrow\textbf{0.07}$)} & \textbf{72.19($\uparrow\textbf{0.63}$)} \\
        
        & & + linear cust & 88.90($\downarrow0.33$) & 87.48($\downarrow0.31$) & 85.24($\downarrow0.34$) & 69.93($\downarrow1.93$) \\
        
        & & + linear basis cust & 88.52($\downarrow0.71$) & 87.64($\downarrow0.15$) & \textbf{85.68($\uparrow\textbf{0.10}$)} & 70.12($\downarrow1.44$) \\

        \cline{2-7}
        
        & Description & + SPhoBert & \textbf{90.11}($\uparrow\textbf{0.88}$) & \textbf{88.06}($\uparrow\textbf{0.27}$) & \textbf{87.22}($\uparrow\textbf{1.64}$) & \textbf{73.49}($\uparrow\textbf{1.93}$) \\

        \cline{2-7}
        & \multirow{2}{*}{Both} & + bias cust & \multirow{2}{*}{89.20($\downarrow0.03$)} & \multirow{2}{*}{\textbf{88.14($\uparrow\textbf{0.35}$)}} & \multirow{2}{*}{\textbf{85.71($\uparrow\textbf{0.13}$)}} & \multirow{2}{*}{\textbf{72.57($\uparrow\textbf{1.01}$)}} \\
        
        & & + SPhoBert &  & & & \\
        
        \hline

        \multirow{8}{*}{BERT4News} & \multicolumn{2}{l|}{Original} & 88.47 & 88.22 & 85.14 & 72.55 \\
        \cline{2-7}
        
         & \multirow{4}{*}{Category} & + bias cust & \textbf{88.80($\uparrow\textbf{0.33}$)} & 87.59($\downarrow0.63$) & \textbf{85.32($\uparrow\textbf{0.18}$)} & 71.58($\downarrow0.97$) \\

        & & + bias basis cust & \textbf{88.57($\uparrow\textbf{0.10}$)} & 87.86($\downarrow0.36$) & 84.68($\downarrow0.46$) & 70.67($\downarrow1.88$) \\
        
        & & + linear cust & \textbf{88.57($\uparrow\textbf{0.10}$)} & 86.07($\downarrow1.52$) & 84.67($\downarrow0.47$) & 66.04($\downarrow6.51$) \\
        
        & & + linear basis cust & 88.42($\downarrow0.05$) & 87.76($\downarrow0.46$) & \textbf{84.63($\uparrow\textbf{0.51}$)} & 70.61($\downarrow1.94$) \\

        \cline{2-7}

        & Description & + SBert4News & \textbf{89.23($\uparrow\textbf{0.76}$)} & 87.51($\downarrow0.71$) & \textbf{85.86($\uparrow\textbf{0.72}$)} & 71.89($\downarrow0.66$) \\

        \cline{2-7}
        & \multirow{2}{*}{Both} & + bias basis cust & \multirow{2}{*}{88.37($\downarrow0.10$)} & \multirow{2}{*}{87.66($\downarrow0.56$)} & \multirow{2}{*}{85.05($\downarrow0.09$)} & \multirow{2}{*}{72.33($\downarrow0.22$)} \\
        
        & & + SBert4News &  & & & \\
        
        \hline
        \end{tabular}}}
    \label{tab:experimental_results}
\end{table}

Next, we evaluated the effectiveness of the classification models when incorporated with the product description. In our experiments, we employed both SPhoBert and SBert4News for the DNN models and we only combined SPhoBert and SBert4News with PhoBERT and BERT4News models, respectively. According to Table \ref{tab:experimental_results}, the DNN models also showed significant improvements when combined with SBert4News. Particularly, BiLSTM exhibited the most notable improvement with F1-macro increasing by 2.98\% and 4.73\% for both tasks. For the BERT models, there were also improvements observed on PhoBERT for both tasks, resulting in F1-macro increases of 1.64\% and 1.93\%, respectively. However, when combined with BERT4News, the accuracy increased for Task 1 (by 0.72\%) but decreased for Task 2 (by 0.66\%). These findings suggest that incorporating the product description into the classifier has the potential to enhance the efficiency of label prediction for reviews.

Lastly, we conducted experiments to incorporate both the category and description of the product into the classifier. The combination types are selected rely on the best results achieved for each attribute. When combining both attributes in DNN models, we observed a significant increase in Accuracy and F1-macro, ranging from approximately 1\% to 4\%. However, it is apparent that this combination does not achieve higher results compared to each attribute, it has rebalanced the accuracy. In the case of the BERT models, the PhoBERT model demonstrated an improvement in F1-macro by 0.13\% and 1.01\% for both tasks, whereas BERT4News experienced a decrease in performance.

In short, based on the aforementioned results, we have observed that metadata has a significant impact on the performance of models. For DNN models, both the product category and product led to improvements in the results, with the incorporation of the category alone achieving optimal results. However, for the BERT models, effectiveness was only enhanced when combining the description and category, while combining them separately proved to be ineffective and resulted in a decrease in model accuracy. This suggests that the BERT models are optimized for learning and understanding sentence semantics, thus benefiting from the inclusion of product description features to better recognize and provide accurate predictions.

\subsection{Error Analysis}

\begin{figure}[H]
    \centering
    \resizebox{\textwidth}{!}{
    \centering
    \begin{minipage}[t]{.4\textwidth}
        \includegraphics[scale=0.4]{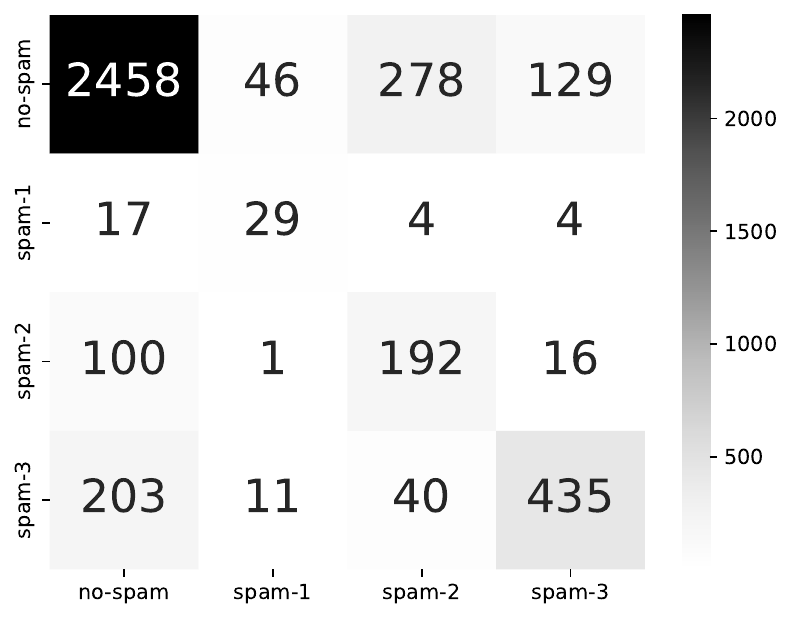}
        \newline \small \centering TextCNN original
    \end{minipage}
    \hspace{.6cm}
    \begin{minipage}[t]{.4\textwidth}
        \includegraphics[scale=0.4]{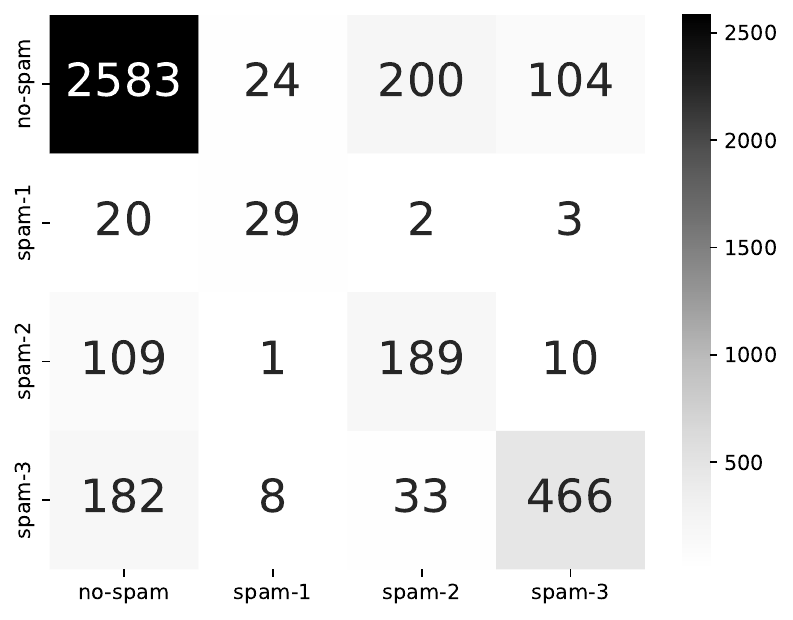}
        \newline \small \centering TextCNN + linear basis cust
    \end{minipage}
    }
    \caption{The confusion matrics of the original TextCNN model and when combining product category on Task 2.}
    \label{fig:confusion_matrix_TextCNN_category_2}
\end{figure}

To evaluate the effectiveness of the integrated metadata proposal method on each label, we analyzed their impact when combining individual attributes and both. Firstly, we chose the model that showed the highest performance improvement when combined with the product category to analyze its efficiency, namely TextCNN + linear basis cust. When comparing the effectiveness of incorporating the product category into the spam reviews classification task, we noticed an improvement in accurately predicting reviews for both labels. To provide a detailed analysis of each spam type, we illustrated the confusion matrix of this model for the task of identifying spam reviews in Figure \ref{fig:confusion_matrix_TextCNN_category_2}. For this task, we observed that the labels NO-SPAM and SPAM-3 both exhibited improved classification results, whereas there was little significant difference observed for the remaining two labels.

To observe the influence of each product category on the prediction results from TextCNN + linear basis cust model compared to the original TextCNN model, we conducted a statistical analysis of misclassified reviews from the original TextCNN model and correctly predicted reviews on these reviews from the TextCNN + linear basis model for the NO-SPAM and SPAM-3 labels, as presented in Table \ref{tab:efficience_of_category}. According to Table \ref{tab:efficience_of_category}, for the NO-SPAM label, the categories that yielded the best results are Vehicle, Home - Life, Fashion, and Beauty. Similarly, for the SPAM-3 label, the categories that showed high effectiveness are Sport, Phone - Computer, Toy, Beauty, Fashion, etc. This observation is also correlated with the statistics of label distribution in each product category, where these categories tend to have higher proportions of the NO-SPAM and SPAM-3 labels compared to the other labels.

\begin{table}[H]
    \renewcommand{\arraystretch}{1.25}
    \centering
    \caption{The effectiveness of integrating product category into NO-SPAM and SPAM-3 labels.}{
    \resizebox{\textwidth}{!}{
    \newcolumntype{P}[1]{>{\centering\arraybackslash}p{#1}}
    \begin{threeparttable}
    \begin{tabular}{|l|P{0.12\textwidth}P{0.13\textwidth}P{0.12\textwidth}|P{0.12\textwidth}P{0.13\textwidth}P{0.12\textwidth}|}
        \hline
        \multirow{2}{*}{\textbf{Category}} & \multicolumn{3}{c|}{\textbf{NO-SPAM}} & \multicolumn{3}{c|}{\textbf{SPAM-3}}  \\
        \cline{2-7}
        & \textbf{Number of wrong} & \textbf{Number of correct} & \textbf{Efficiency (\%)} & \textbf{Number of wrong} & \textbf{Number of correct} & \textbf{Efficiency (\%)} \\
        \hline
        Fashion & 34 & 19 & 55.88 & 57 & 12 & 21.05 \\
        Electronic & 75 & 27 & 36.00 & 15 & 2 & 13.33 \\
        Others & 73 & 33 & 45.21 & 38 & 8 & 21.05 \\
        Phone - Computer & 51 & 19 & 37.25 & 28 & 9 & 32.14 \\
        Beauty & 37 & 19 & 51.35 & 20 & 6 & 30.00 \\
        Home - Life & 44 & 26 & 59.09 & 24 & 4 & 16.67 \\
        Toy & 40 & 14 & 35.00 & 13 & 4 & 30.77 \\
        Sport & 21 & 7 & 33.33 & 19 & 7 & 36.84 \\
        Camera & 34 & 17 & 50.00 & 7 & 1 & 14.29 \\
        Vehicel & 22 & 17 & 77.27 & 15 & 4 & 26.67 \\
        Footwear & 5 & 0 & 00.00 & 11 & 2 & 18.18 \\
        Book & 17 & 8 & 47.06 & 7 & 0 & 00.00 \\
        \hline
        Total & 453 & 206 & 45.47 & 254 & 59 & 23.23 \\
        \hline
    \end{tabular}
    \vspace{2pt}
    \begin{tablenotes}[para,flushleft]
         \smallskip\item\hspace{-2.5pt}\noindent\textbf{Number of wrong} is the count of incorrectly predicted reviews from the original TextCNN model for the two labels: NO-SPAM and SPAM-3. \textbf{Number of correct} is the count of reviews correctly predicted by the TextCNN + linear basis cust model on the previously misclassified reviews.
        \end{tablenotes}
    \end{threeparttable}
    }}
    \label{tab:efficience_of_category}
\end{table}

Furthermore, we also provided several examples of reviews with the NO-SPAM and SPAM-3 labels, where the TextCNN model misclassified them but the TextCNN + linear basis cust model predicted them correctly, as illustrated in Table \ref{tab:result_textcnn_basis_linear_cust}. We can see that reviews \#1 and \#2 belonged to the NO-SPAM label because the content of the reviews was related to the products being sold. However, they mentioned the Vinfast (a Vietnamese car brand) or Spirituality \& Science topic (a book genre), resulting in misclassification as SPAM-2, contrary to the actual label and after incorporating the product categories, the model accurately predicted the correct label. Conversely, reviews \#3, \#4, and \#5 belonged to the SPAM-3 label due to the presence of spam words in their content. However, the TextCNN model incorrectly predicted them as the NO-SPAM label and upon incorporating the product categories, the model also provided accurate predictions.

\begin{table}[H]
    \definecolor{mayablue}{rgb}{0.45, 0.76, 0.98}
    \renewcommand{\arraystretch}{1.25}
    \caption{Some examples of reviews predicted by the TextCNN and TextCNN + linear basis cust models.}{
    \resizebox{\textwidth}{!}{
    \newcolumntype{P}[1]{>{\centering\arraybackslash}p{#1}}
    \begin{tabular}{cp{0.75\textwidth}P{0.1\textwidth}ccP{0.2\textwidth}}
         \hline
         \multirow{2}{*}{\textbf{No.}} & \multirow{2}{*}{\textbf{Comment}} & \multirow{2}{*}{\textbf{Category}} & \multirow{2}{*}{\textbf{Actual Label}} & \multicolumn{2}{c}{\textbf{Predicted Label}} \\
         \cline{5-6}
         & & & & \textbf{TextCNN} & \textbf{TextCNN + linear basis cust} \\
         \hline
         1 & \textbf{Vietnamese:} Mình mua trong đợt khuyến mãi tặng pin, nên vừa dc xe chất lượng vừa rẻ luôn. Thích lắm. Mình rất khi viết đánh giá sản phẩm đã mua, nhưng vì quá ưng ý nên cũng ráng dành thời gian nhận xét, cũng như động viên công ty \colorbox{mayablue}{Vinfast} phát triển thêm nhiều sản phẩm thân thiện môi trường. \newline \textbf{English:} I purchased the product during a promotional period that included a free battery, so not only did got a high-quality vehicle, but it was also at a very affordable price. I like it. I hardly write reviews for products I've purchased, but because I am extremely satisfied, I take the time to provide feedback, as well as to encourage \colorbox{mayablue}{Vinfast} company  to continue developing more environmentally friendly products. & Vehicle & NO-SPAM & SPAM-2 & NO-SPAM
         \\
         \hline
         
         2 & \textbf{Vietnamese:} Shop chuẩn bị hàng K lâu vì quá Lâu. cơ mà sách bọc kĩ nên sẽ tha thứ. nội dung sách hay liên quan đên \colorbox{mayablue}{Tâm Linh \& Khoa Học}.Đúng cái mk cần. sách đẹp, thơm. Có lẽ sẽ đặt thêm 1 cuốn sách bìa mềm nữa để dự phòg. Sách sau này k SX nữa. để con cháu sau này nó còn mở mang tầm mắt =)) sorry vì đi quá xa \newline \textbf{English:} The shop took too long to prepare the order, but the book was well-packaged, so I can forgive that. The content of the book is interesting and relates to \colorbox{mayablue}{Spirituality \& Science}, which is exactly what I was looking for. The book is beautifully designed and has a pleasant scent. Perhaps I will order an additional softcover book as a backup, as it seems that this book will no longer be produced. This to future generations can also broaden their horizons :)) sorry for going off-topic. & Book & NO-SPAM & SPAM-2 & NO-SPAM \\
         \hline
         
         3 & \textbf{Vietnamese:} Đp nói nhiều xà ngon vcl \colorbox{pink}{đjksjcjskahfjsiajcjfjfjdjfjfjdjdjdjfjdjfjfnfjjfjfj} \newline \textbf{English:} No need to say anything more, barbell is good \colorbox{pink}{đjksjcjskahfjsiajcjfjfjdjfjfjdjdjdjfjdjfjfnfjjfjfj} & Sport & SPAM-3 & NO-SPAM & SPAM-3 \\
         \hline
         
         4 & \textbf{Vietnamese:} \colorbox{pink}{Jsjssjdjdjdjdjsksiiejdjquaanf} rất đẹp và giá hợp lí em rất rất rất \colorbox{pink}{tráhhdhdhdjdjdjkdkdk} thích \colorbox{pink}{luônnn} đánh giá chỉ mang tính chất nhận xu nên mọi người thông cảm, nhưng quần chất lượng thật sự luôn á . Sẽ ủng hộ \colorbox{pink}{thêmmmmmmmmmmmmmmm} \newline \textbf{English:} \colorbox{pink}{Jsjssjdjdjdjdjsksiiejdjpants} are very beautiful and reasonably priced. I really, really, really love \colorbox{pink}{themmm}. Please understand that this review is posted to earn coins, but the quality of the pants is truly outstanding. I will continue to support them. I will continue to support in the \colorbox{pink}{futureeeeeeeeeeeeee}. & Fashion & SPAM-3 & NO-SPAM & SPAM-3 \\
         \hline
         
         5 & \textbf{Vietnamese:} Quạt đẹp ưng ạ \colorbox{pink}{pưlam} \colorbox{pink}{Ksjsjskkskdjxncjosksnxnksksnxbsjsksn} \colorbox{pink}{xnxkxms} \colorbox{pink}{xnxkxmxmx} \colorbox{pink}{xnnxkxnxnxkxm}
         \newline \textbf{English:} The fan is beautiful I am satisfied \colorbox{pink}{pưlam} \colorbox{pink}{Ksjsjskkskdjxncjosksnxnksksnxbsjsksn} \colorbox{pink}{xnxkxms} \colorbox{pink}{xnxkxmxmx} \colorbox{pink}{xnnxkxnxnxkxm} & Electronic & SPAM-3 & SPAM-2 & SPAM-3 \\
         \hline
    \end{tabular}}}
    \label{tab:result_textcnn_basis_linear_cust}
\end{table}

\begin{figure}[H]
    \centering
    \resizebox{\textwidth}{!}{
    \centering
    \begin{minipage}[t]{.45\textwidth}
        \includegraphics[scale=0.45]{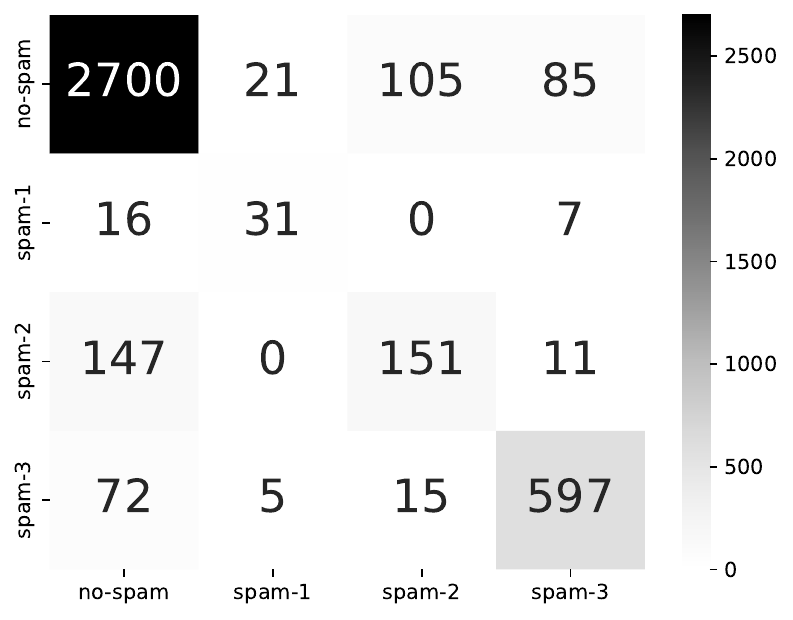}
        \newline \small \centering PhoBERT original
    \end{minipage}
    \hspace{.6cm}
    \begin{minipage}[t]{.45\textwidth}
        \includegraphics[scale=0.45]{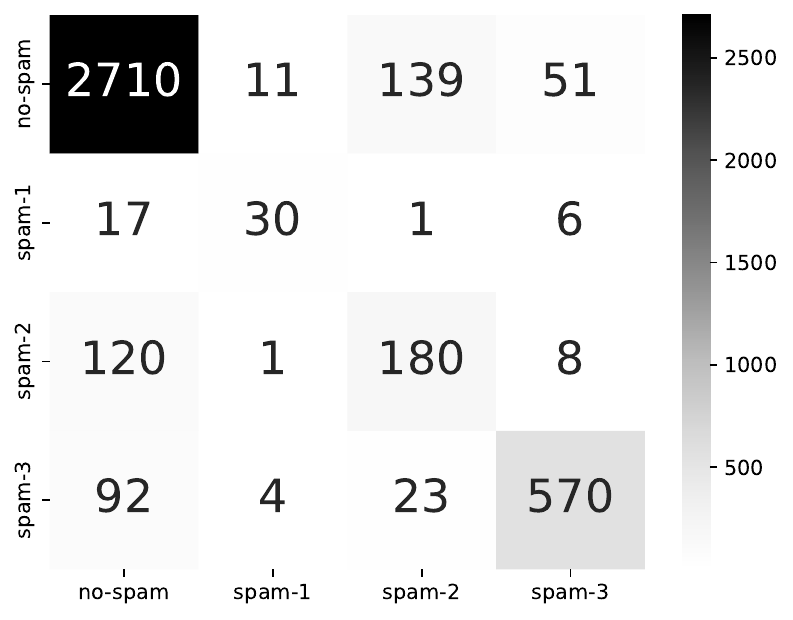}
        \newline \small \centering PhoBERT + SPhoBert
    \end{minipage}
    }
    \caption{The confusion matrics of the original PhoBERT model and when combining feature product description on Task 2.}
    \label{fig:confusion_matrix_PhoBERT_description_2}
\end{figure}

Next, we evaluate the performance of the product description integration method, we compared the model that achieved the highest results when combining this attribute, namely PhoBERT and the combination of features generated from SPhoBert (PhoBERT + SPhoBert). The confusion matrices of these two models for Task 2 are illustrated in Figure \ref{fig:confusion_matrix_PhoBERT_description_2}. According to Figure \ref{fig:confusion_matrix_PhoBERT_description_2}, it is evident that the number of accurately predicted reviews with the SPAM-2 labels has increased from 151 to 180, and there have been marginal increases in misclassification rates for the remaining labels. The SPAM-2 label poses a challenge for the baseline model, and the integration of product descriptions has improved the accuracy of predictions for this particular label. This demonstrates that the product description contains valuable information that aids the model in better-identifying types of spam reviews.

Additionally, we compared the prediction results of some reviews from the PhoBERT and PhoBERT + SPhoBert models, as shown in Table \ref{tab:result_phobert_sphobert}. According to Table \ref{tab:result_phobert_sphobert}, most of the reviews that only focused on the brand, such as \#1, \#2, \#3, \#5, and \#6, yielded accurate predictions when product descriptions were incorporated. Moreover, it is worth noting that even brief reviews such as \#3 and \#8 achieved accurate prediction outcomes. Only review \#7 is still predicted incorrectly, possibly because it is evaluated based on the brand but contains the phrase "will support further" which is commonly found in NO-SPAM reviews, causing the model to overlook it. However, the effectiveness of the product description feature in enhancing the prediction accuracy of type 2 spam reviews is clearly evident.

Finally, we compare the results when combining both product category and description, although the overall improvement in accuracy was not higher compared to incorporating them individually, each label showed better prediction results. From the results of the error analysis, we conclude that metadata helps enhance the effectiveness of classifying spam reviews, depending on specific cases, we can combine either attribute or both. If expecting better results for the SPAM-2 labels, combining product descriptions would be effective. If aiming for high performance in NO-SPAM and SPAM-3 labels, combining product categories can be employed. If a balanced improvement across labels is desired, combining both product categories and descriptions would be suitable.

\begin{table}[H]
    \definecolor{mayablue}{rgb}{0.45, 0.76, 0.98}
    \renewcommand{\arraystretch}{1.25}
    \centering
    \caption{Some reviews are predicted by PhoBERT and PhoBERT + SPhoBert models.}{
    \resizebox{\textwidth}{!}{
    \begin{tabular}{cp{10cm}cccc}
         \hline
         \multirow{2}{*}{\textbf{No.}} & \multirow{2}{*}{\textbf{Comment}} & \multirow{2}{*}{\textbf{Actual Label}} & \multicolumn{2}{c}{\textbf{Predicted Label}} \\
         \cline{4-5}
         & & & \textbf{PhoBERT} & \textbf{PhoBERT + SPhoBert} \\
         \hline
         1 & \textbf{Vietnamese:} "Cảm ơn \colorbox{mayablue}{Tefal} về chất lượng sản phẩm và cảm ơn \colorbox{mayablue}{Tiki} về chất lượng dịch vụ. Chất lượng sản phẩm \colorbox{mayablue}{Tefal} luôn vượt qua sự mong đợi của mình". \newline \textbf{English:} "Thank you \colorbox{mayablue}{Tefal} for product quality and thank you \colorbox{mayablue}{Tiki} for service quality. \colorbox{mayablue}{Tefal} product quality always exceeds my expectations". & SPAM-2 & NO-SPAM & SPAM-2 \\
         \hline
         2 & \textbf{Vietnamese:} Hàng \colorbox{mayablue}{Xiaomi} luôn ngon, bổ, rẻ. Cám ơn, luôn ủng hộ \colorbox{mayablue}{Xiaomi} và \colorbox{mayablue}{Tiki} \newline \textbf{English:} \colorbox{mayablue}{Xiaomi} products are always great, reliable, and affordable. Thank you, always support \colorbox{mayablue}{Xiaomi} and \colorbox{mayablue}{Tiki}.. & SPAM-2 & NO-SPAM & SPAM-2 \\
         \hline
         3 & \textbf{Vietnamese:} giá ok, giao hàng nhanh, đóng gói kỹ \newline \textbf{English:} the price is good, fast delivery, and careful packaging & SPAM-2 & NO-SPAM & SPAM-2 \\
         \hline
         4 & \textbf{Vietnamese:} \colorbox{mayablue}{Philips} thì miễn bàn về chất lượng rồi! Thời gian giao hành bên \colorbox{mayablue}{Tiki} cũng nhanh, hài lòng 100\% \newline \textbf{English:} \colorbox{mayablue}{Philips} is known for its excellent quality! The delivery time with \colorbox{mayablue}{Tiki} is also fast, and I am extremely satisfied. & SPAM-2 & NO-SPAM & SPAM-2 \\
         \hline
         5 & \textbf{Vietnamese:} \colorbox{mayablue}{Pop It} là đồ chơi xả stress theo lời con gái khoe. Shop giao hàng nhanh, trao đổi nhiệt tình \newline \textbf{English}: \colorbox{mayablue}{Pop It} is a stress reliever according to what the girl said. Shop delivered products quickly, enthusiastic exchange & SPAM-2 & NO-SPAM & SPAM-2 \\
         \hline
         6 & \textbf{Vietnamese:} \colorbox{mayablue}{Cafe Highland} chất lượng thì không bàn cãi rồi. Đóng gói cẩn thận. Giao hàng nhanh chóng. \newline \textbf{English}: The quality of \colorbox{mayablue}{Highland coffee} is beyond question. The packaging is careful. The delivery is fast. & SPAM-2 & NO-SPAM & SPAM-2 \\
         \hline
         7 & \textbf{Vietnamese:} mình mua ở \colorbox{mayablue}{US247 Mart} tin tưởng sẽ ủng hộ thêm \newline \textbf{English}: I purchased it at \colorbox{mayablue}{US247 Mart}, I trust and will support further. & SPAM-2 & NO-SPAM & NO-SPAM \\
         \hline
         8 & \textbf{Vietnamese:} giao hàng chậm \newline \textbf{English}: the delivery is slow. & SPAM-2 & NO-SPAM & SPAM-2 \\
         \hline
    \end{tabular}}}
    \label{tab:result_phobert_sphobert}
\end{table}

%% file: conclusion.tex
\section{Conclusion}
\label{conclusion}

In this work, we have introduced the ViSpamReviews v2 corpus for the purpose of integrating metadata for detecting spam reviews on Vietnamese e-commerce websites. We have proposed an approach to incorporate metadata into this problem, including two attributes: product category and product description. Through experiments and evaluations, the results show that the PhoBERT model combined with the product description feature generated from SPhoBert achieves the highest performance in two tasks: spam review classification and spam review type identification, with F1-macro scores of 87.22\% (an increase of 1.64\%) and 73.49\% (an increase of 1.93\%), respectively. This demonstrates the potential of metadata in enhancing the predictive capabilities of classification models for the spam review detection task.

Although the results have shown improved predictive efficacy on different labels, the accuracy of the models still has limitations. Upon observing some reviews, we noticed that the rule-based method can perform well in identifying the SPAM-3, and rule sets can be constructed to pre-classify such reviews. Additionally, user comments in reviews often contain many abbreviations, which partly complicates the learning process for the models. Therefore, it is necessary to build a dictionary for standardizing these abbreviations. Finally, we propose combining both rule-based methods and training models on normalized data, along with researching and experimenting with methods to better leverage metadata information to enhance the performance of the models.